# Performance Evaluation of Machine Learning Algorithms in Post-operative Life Expectancy in the Lung Cancer Patients

Kwetishe Joro Danjuma[1]

[1] Department of Computer Science, Modibbo Adama University of Technology,
Yola, Adamawa State, Nigeria

**Abstract**

The nature of clinical data makes it difficult to quickly select, tune and apply machine learning algorithms to clinical prognosis. As a result, a lot of time is spent searching for the most appropriate machine learning algorithms applicable in clinical prognosis that contains either binary-valued or multi-valued attributes. The study set out to identify and evaluate the performance of machine learning classification schemes applied in clinical prognosis of post-operative life expectancy in the lung cancer patients. Multilayer Perceptron, J48, and the Naive Bayes algorithms were used to train and test models on Thoracic Surgery datasets obtained from the University of California Irvine machine learning repository. Stratified 10-fold cross-validation was used to evaluate baseline performance accuracy of the classifiers. The comparative analysis shows that multilayer perceptron performed best with classification accuracy of 82.3%, J48 came out second with classification accuracy of 81.8%, and Naive Bayes came out the worst with classification accuracy of 74.4%. The quality and outcome of the chosen machine learning algorithms depends on the ingenuity of the clinical miner.

## 1. Introduction

In clinical medicine, time plays crucial role in disease prognosis as well as data collection and decision-making. The healthcare system generates an unprecedented terabytes of data leading to information overload, and the ability to make sense of such data is becoming increasingly important with in-depth knowledge of exploratory data analysis and machine learning scheme [1]. The healthcare system is data rich, information poor. The system generates unprecedented volume of data, but lack effective analysis tools to extract and discover hidden knowledge. It is almost impossible to make sense of very large data without appropriate computer programs such as Spreadsheet, data visualization software, statistical packages, OLAP (Online Analytical Processing) application, and data mining [1]. The advances in data collection and processing require new techniques and tools to intelligently transform the unprecedented volume of data into useful information [2] that support clinical prognosis and patient care. Data mining is a process of nontrivial extraction of implicit, previously unknown and potentially useful information from the data stored in a database [3]. Data mining finds correlation and patterns among attributes in a very large datasets to build up knowledgebase based on the given constraint. The knowledge extraction, transformation and representation in human understandable structure for further use are often referred to as Knowledge Discovery in Databases [4], which deals with inconclusive, noisy and sparse data to finding valid, useful, novel and understandable patterns in data [2]. The concept of knowledge discovery in databases (KDD) encompasses data storage and access, and scaling machine learning algorithms to very large datasets and interpreting the results [4]. KDD also involves different data mining algorithms used to build models that enable unknown data to learn to identify new information. The most commonly associated feature of data mining techniques regardless of origin is the automated discovery of relationships and dependencies of attributes in the observed data [5]. The automated discovery of relationships is supported by many machine learning algorithms such as Artificial Neural Networks (ANN), Cluster Analysis (CA), Genetic Algorithms (GA), Support Vector Machines (SVM), and Decision Trees (DTs) to predict future trends and behaviours, allowing businesses to make proactive, knowledge-driven decisions [6]. Though C4.5, k-means, SVM, Apriori, EM (expectation maximization), PageRank, AdaBoost, kNN (k-nearest neighbours), Naïve Bayes, and Classification and Regression Tree (CART) have been identified as the most influential algorithms for classification, clustering, regression, association rules, and network analysis ranked based on expert nominations, citation counts, and a community survey [7].

Predictive analytics comprises of machine learning algorithms such as artificial neural networks (ANN) and decision trees (DTs) among a myriad of other algorithms used in knowledge extractions, and apply the obtained knowledge to detect or predict trends in new data [8]. The widespread availability of new computational methods and tools for data analysis and predictive modelling requires medical informatics and clinicians to systematically select the most appropriate strategy to cope with clinical prediction problems [9]. The machine learning process support clinicians and medical informatics to analyse retrospective data, and to exploit large amount of data routinely collected in their day-to-day activity [10]. The data is analysed to extract useful information that supports disease prognosis and to develop models that predict patient's health more accurately [6]. Data mining can be used in a predictive manner in a variety of application for fraud and intrusion detection, market basket analysis (MBA), customer segmentation and marketing, phenomena of "beer & baby diapers", corporate surveillance and criminal investigation, financial and risk management, and medical and healthcare [11]. In this paper, we set out to identify and evaluate the performance of machine learning classification schemes applied in the prediction of post-operative life expectancy in the Lung Cancer patients.

## 2. Literature Review

### 2.1 Application of Data Mining in Clinical Medicine

In clinical diagnosis and prognosis, machine learning classification schemes are classified into three categories: those used for disease diagnosis, disease prognosis, or both diagnosis and prognosis [3]. Clinical prognosis encompasses the science of estimating the complication and recurrence of disease and to predict the survival of patient or group of patients. In other words, it involves prediction modelling estimation of different parameters related to patient's health. Survival analysis applies various techniques to estimate the survival of a particular patient suffering from disease over a particular time period, defined as a patient remaining alive for a specified time period of 10 years or longer after the disease diagnosis. Unfortunately, survival estimates developed using such a definition of survival may not accurately reflect the current state of treatment and the probability of survival. However, improvement in early detection and treatment will increase the expectations of survival [12]. The most influential of these data mining techniques for classification, clustering, regression, association rules, and network analysis ranked based on expert nominations, citation counts and community survey were identified in the work of [7] to include C4.5, k-means, Support Vector Machine (SVM), Apriori, Expectation Maximization (EM), PageRank, Adaboost, k-nearest neighbor (kNN), Naïve Bayes, and CART.

### 2.2 Application of Predictive Data Mining in Clinical Prognosis

In disease prognosis, [13] examined potential use of classification based data mining techniques such as Rule based DT, Naïve Bayes and ANN in the prediction of heart attack. In an analysis of cancer data in building prediction models for prostate cancer survivability, [14] used DT, ANN and SVM alongside logistic regression to develop prediction models for prostate cancer survivability. A k-fold cross-validation methodology was used in model building, evaluation and comparison, and SVM performed best followed by artificial neural networks and decision trees. In the investigation of expected survival time of patients with pancreatic cancer, [15] demonstrated that machine learning algorithms such as ANN, Bayesian Networks, and SVM are capable of improved prognostic predictions of pancreatic cancer patient survival as compared with logistic regression alone. A review of 349 patients who underwent ACL reconstruction at outpatient surgical facility, [16] developed machine learning classifiers based on logistic regression, BayesNet, Multilayer perceptron, SVM, and Alternating decision tree (ADTree) to predict which patients would require postoperative Femoral Nerve Block (FNB). The Machine Learning algorithms specifically the ADTree outperformed traditional logistic regression with regards to Receiver Operating Curve (ROC), and vice-versa with regard to kappa statistics and percent correctly classified. In prognosis of tumor detection and classification in digital mammography, [17] applied back-propagation neural networks and constraint form association rule mining for tumor classification in mammograms. The formal proved to be less sensitive at a cost of high training times compared to the latter with better results than reported in related literatures.

In prediction of breast cancer prognosis, [18] developed an ANN, a Bayesian network, and a hybrid Bayesian network (HBN) that combined ANN and Bayesian network to obtain a good estimation of prognosis as well as a good explanation of the results. The HBN and ANN models outperformed the Bayesian network model in breast cancer prognosis. [19] Presented a comparative analysis of the Naïve Bayes, the back-propagated neural network, and the C4.5 decision tree algorithms in prediction of breast cancer survivability rate. The C4.5 algorithm had a much better performance compared to others. In cancer prognosis, [20] investigated a hybrid scheme based on fuzzy decision trees for cancer prognosis. Performance comparisons suggest hybrid fuzzy decision tree classification is more robust and balanced than independently applied crisp classification. A comparative analysis of algorithms show that DTs, ANNs and Bayesian are the well-performing algorithms used for disease diagnosis, while ANNs is the well-performing algorithm, followed by Bayesian, DTs and Fuzzy algorithms.

## 2.3 Uniqueness of Clinical Data Mining and Ethics

Clinical researchers from other disciplines are often unaware of the particular constraints and difficulties involved in mining privacy-sensitive, heterogeneous, and voluminous clinical data. The miner has the responsibility to conduct valid research in a manner that ensures patient confidentiality by anonymizing data. The select of medicine as a researcher resource should be secondary to patient-care activity [21]. The ethical, security and legal aspects of clinical data mining includes data ownership, fear of lawsuits, expected benefits, and special administrative issues [22]. Patient medical records almost always encompass patient's age, diseases suffer or suffered from, and whether they smoke cigarettes or not, are often elicited to serves as basis for preliminary diagnosis and or prognosis. Clinical data often originate from clinical consultation with patients, medical images, ECG, EEG and RTG signals, physician's notes and interpretations and other screening results that may bear upon clinical analysis and treatment of the patient [22-23]. The data may contain noise, contradiction, missing values and important information (signs, symptoms, clinical reasoning, and so on) that may be stored in an unstructured way. However, the assessment of clinical data quality, data coding quality, unstructured data transformation to structure aid detection of medical errors that bears upon clinical prognosis. A much better approach to deal with clinical data have been developed to support the retrieval of narrative or imaging documents, to classify medical reports automatically, and to preserve the patient's privacy and confidentiality in medical reports for secondary usage [24].

Often time, clinical appointment is gathered in decision table with conditional and decisional attributes. The conditional attribute is defined as the set $S = \{s_1, s_2, ..., s_i\}$ where $S$ is symptoms, and the decisional attribute is defined as the set $D = \{d_1, d_2, ..., d_k\}$ where $D$ represent diseases. If $P = \{p_1, p_2, ..., p_N\}$ is defined as the set of patients, then the decision table can be constructed as the set of quadruple in Eq. (1)

$$T = \{P, S, D, p\} \quad (1)$$

Where $\rho$ is a function represented in Eq. (2)

$$p \bullet P*(S \cup D) \rightarrow \{w_{dk}\} \quad (2)$$

The values of symptoms are marked with the symbol $v_{i,n}$, which denote a symptom value for $i-th$ symptom and $n-th$ patient. The values of diseases are marked with $w_{k,dk}$ for $k-th$ disease and $d_k - th$ value. The values of $v_{i,n}$ are usually binary with 1 denotes occurrence of symptoms and 0 denotes lack of occurrence. Often time clinical data are positive-valued (except in the case of ECG), and the values of symptoms ordinarily fit into definite range. For instance resting blood pressure is no lower than 30 and no higher than 300 [23].

## 2.4 Handling Imbalanced Clinical Data

Clinical data are often noisy and incomplete resulting in missing values. In clinical prognosis, it is difficult to specify a likely range of values to replace missing values without biasedness [23]. The missing values could be substituted with a) most likely values; b) all possible values for that attribute; and c) intermediate approach of specifying a likely range of values in an unbiased manner, instead of only one most likely [22]. Another simple method to handle missing values imputation includes; a) data reduction and elimination of all missing values; b) use of most common values, mean or median; and c) closest fit approach and methods based on machine learning algorithms such as k-nearest neighbor, neural networks and association rules [25]. A dataset is said to be imbalanced when the number of instances of one class is much lower than the instances of the other class, often referred to as "rare classes" [26], or when the classification categories are not approximately equally represented due to class distribution or costs of errors instances. Resampling techniques such as random sampling with replacement, random under-sampling, focused oversampling, focused under-sampling, oversampling with synthetic generation of new samples based on the known information, and combinations of the above techniques have been used in handling imbalanced data [27]. Data imbalance can be resolved either at: a) data level using either oversampling the minority class (Positive instances) or under-sampling the majority class (Negative instances) or both; or b) algorithmic level adjusting classifier to be trained by modifying the class cost, establishing a bias towards the positive class, learning from just one class (recognition based learning) instead of learning from two classes (discrimination based learning) [26].

Sampling techniques such as under-sampling the majority class, oversampling the minority class, changing score-based classifier decision threshold, and modifying algorithms to incorporate different weights for errors on positive and negative instances are used to handle imbalanced data [28]. In an attempt to re-balance imbalanced datasets, [29] proposed the use of: a) under-sampling of the majority class; b) over-sampling of the minority class; c) modify the sensitivity of the classifier so that errors on minority class (positive), to be costlier than errors on the majority class (negative); and d) Synthetic Minority Oversampling Techniques (SMOTE). SMOTE creates new synthetic training data for the minority class by adding random value (SMOTE-Randomize) to some features of the original training data, and providing new data that lies close to the original ones in the multi-dimensional space of the problem. Support vector machines (SVM) in which asymmetrical margins are tuned to improve recognition of rare positive cases; and a new resampling approach in which both oversampling and under-sampling rely on synthetic cases (prototypes) generated through class specific sub-clustering are used in solving data imbalance problem [30]. [31] Proposed a hybrid sampling technique which incorporates both over-sampling and under-sampling, with an ensemble of support vector machines (SVM) for learning from imbalanced data to improve the prediction performance.

## 3. Predictive Data Mining Algorithms to Compare

3.1 Decision Tree Algorithm

Decision Tree is a popular classifier in machine learning environment that is simple and easy to implement, and requires no domain knowledge or parameter setting and can handle high dimensional data, more appropriate for exploratory knowledge discovery and analysis [32-33]. Decision tree is a non-parametric supervised learning algorithms that model non-linear relations between predictors and outcomes and for mixed data types (numerical and categorical), isolates outliers, and incorporates a pruning process using cross-validation as an alternative to testing for unbiasedness with a second data set [34]. It predicts the value of a target variable by learning simple decision rules inferred from the data features [35]. It is the most widely used machine learning algorithms in clinical prognosis capable of handling continuous attributes that are essential in case of medical data [23]. The decision tree is popular and widely used because of its shorter learning curve and interpretability, and the tree ability to handle covariates attributes measured at different level [36]. Decision trees are significantly faster than neural networks with a shorter learning curve that is mainly used in the classification and prediction to represent knowledge. The instances are classified by sorting them down the tree from the root node to some leaf node. The nodes are branching based on if-then condition [37].

There are many variants of decision tree such as CART, ID3, C4.5, SLIQ, and SPRINT [38]. The classification and regression tree (CART), Chi-squared automatic interaction detector (CHAID), quick-unbiased efficient statistical tree (QUEST), C4.5 and Interactive Dichotomiser (ID3) are more suitable than classical statistical methods. It uses recursive partitioning to assess the effect of specific variables on survival, thereby generating groups of patients with similar clinical features and survival times in a tree-structured model that can be analyzed to assess its clinical utility [34] & [5]. Also, decision trees are praised for their transparency, allowing bioinformatics experts to examine and understand the decision model and its workings, and each path in the decision tree can be regarded as a decision rule [9]. The decision tree is built of nodes which specify conditional attributes – symptoms $S = \{s_1, s_2, ..., s_i\}$, branches which show the values of $v_{i,k}$ i.e. the $h-th$ range for $i-th$ symptom and leaves which present decisions $D = \{d_1, d_2, ..., d_k\}$ and their binary values, $w_{dk} = \{0,1\}$. A decision tree may be converted to a set of association rules by writing down each path from the root to the leaves in a form of rules [23]. For instance, the decision tree in Fig.1 can be written as a set of association rule in Eq. (3).

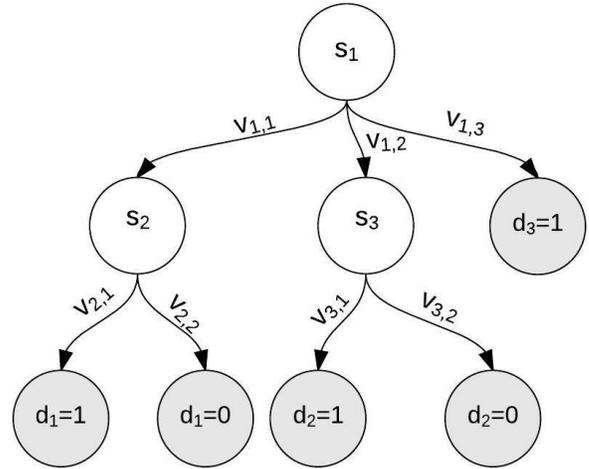

Fig.1 Sample decision tree applicable in Clinical Prognosis

$$(S_1, v_{1,1}) \cap (S_2, v_{2,1}) \Rightarrow (d_1 = 1)$$
$$(S_1, v_{1,1}) \cap (S_2, v_{2,2}) \Rightarrow (d_1 = 0)$$
$$(S_1, v_{1,2}) \cap (S_3, v_{3,1}) \Rightarrow (d_2 = 1)$$
$$(S_1, v_{1,2}) \cap (S_3, v_{3,2}) \Rightarrow (d_2 = 0)$$
$$(S_1, v_{1,3}) \Rightarrow (d_3 = 1) \qquad (3)$$

## 3.2 Naïve Bayes Algorithm

Bayesian network has been successfully applied in diagnosis and antibiotic treatment of pneumonia, and Naïve Bayes algorithm performance has been tested against a colorectal cancer. It takes domain experts and structure learning such as genetic algorithm (GA) to successfully construct network topology from training data [23] & [18]. The principle of Naïve Bayes is based on Bayes rules of simple conditional probability used to estimate the likelihood of a property given small amount of training data to estimate parameters such as mean and variance necessary for classification [39]. Naïve Bayes is a statistical classifier which assumes no dependency between attributes but attempts to maximize the posterior probability in determining the class. The performance of Naïve Bayes has been observed to be consistent before and after attributes reduction [32]. The probabilities applied in the Naïve Bayes algorithm is intended to learn the probability of the likelihood of some symptom $S$ with the highest posterior probability distribution, given some observation or prognosis $x$ where there exist a dependence relationship between $S$ and $x$, denoted as $P(S|x)$, and the posterior probability distribution can be computed as shown in Eq. (4).

$$p(S|x) = \frac{p(x|S)p(S)}{p(x)} \alpha p(x|S)p(S) \quad (4)$$

The posterior probability distribution is proportional to the product of two terms; the marginal likelihood $p(x|S)$ with respect to $x$, and the prior probability of the symptom $p(S)$. Naïve Bayes model forms a network of nodes that are interconnected with directed edges and form a directed acyclic graph [23], used to model the dependencies among variables [15]. Each node in the directed acyclic graph represents a stochastic variable and arcs represent a probabilistic dependency between a node and its parents [9].

## 3.3 Artificial Neural Networks (ANN) Algorithm

ANN models are based on a set of multilayered interconnected equation which uses non-linear statistical analysis to reveal previously unrecognized relations between given input variables and an output variable. It has been found accurate and reliable in disease diagnosis and prediction outcome in diverse clinical trials, by means of symptoms routinely available to clinicians. It has also been found promising for studying neurodegenerative disorders [40]. Despite the complexity and difficulties involved in understanding ANN's prediction, it has been successfully applied in clinical prognosis especially in the phase of coronary artery disease prediction, EEG signals processing and the development of novel antidepressants [23]. ANN is a computational model that is biologically inspired, highly sophisticated analytical techniques, capable of modelling extremely complex non-linear functions [41]. It is a network of highly interconnected processing neurons inspired by biological nervous systems operating in parallel through a subgroup of processing element known as layer in the network. It consists of Input layer, the hidden layer and the output layer, which are trained to perform specific functions by adjusting the values of the weights between elements [42-43]. ANN model is formed by an input layer, one or more hidden layers, and the output layer. The number of neurons and layers depends on the complexity of the system being studied. The neurons in the input layer receive the data, transfer the data to neurons in the hidden layers through the weighted links for processing, and the result is transferred to the neurons in the output layer for analysis [44]. ANN are systems modelled based on the cognitive learning process and the neurological functions of the human brain, consisting of millions of neurons interconnected by synapses [45], and capable of predicting new observations after learning from existing data. As the human brain is capable to, after the learning process, draw assumptions based on previous observations, neural networks are also capable to predict changes and events in the system after the process of learning [37]. The interconnected sets of neurons are divided into three: input, hidden, and output ones.

In clinical medicine, the patient's symptoms could serve as input set $S$, and disease could serve as output set $D$ to the neural network. The hidden neuron processes the outcome of preceding layers. The process of learning in ANN is to solve a task $T$, having a set of observations and a class of functions $F$, which is to find $f^* \in F$ as the optimal solution to the task [23]. The most popular of the ANN is Multilayer Perceptron algorithm (MLP). MLP is most suitable for approximating a classification function, and consists of a set of sensory elements that make up the input layer, one or more hidden layers of processing elements, and the output layer of the processing elements [5]. The Multi-Layer Perceptron (MLP) with back-propagation (a supervised learning algorithm) is arguably the most commonly used and well-studied ANN architecture capable of learning arbitrarily complex nonlinear functions to arbitrary accuracy levels [46], and its ability to process complex problems which a single hidden layer neural network cannot solve [47]. It is essentially the collection of nonlinear neurons (perceptron's) organized and connected to each other in a feed forward multi-layer structure. **Error! Reference source not found.** presents a graphical depiction of ANN for clinical prognosis.

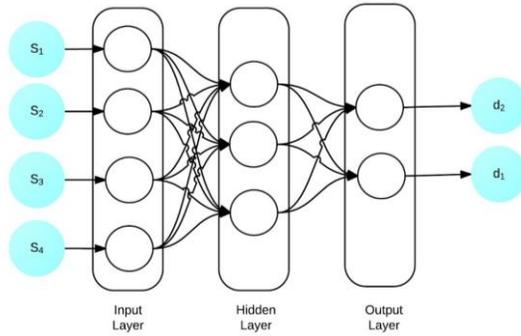

Fig.2 Graphical depiction of ANN for clinical prognosis

## 4. Model Evaluation Metrics

Though empirical studies have shown that it is difficult to decide which metric to use for different problem, each of them has specific features that measure various aspects of the algorithms being evaluated [48]. It is often difficult to state which metrics is the most suitable to evaluate algorithms in clinical medicine due to large weighted discrepancies that often arise between predicted and actual value or otherwise [23]. The performance evaluation of machine learning algorithms is assessed based on predictive accuracy, which is often inappropriate in case of imbalanced data and error costs vary remarkably [27]. Machine learning performance evaluations involve certain level of trade-off between true positive and true negative rate, and between recall and precision. Precision, Recall and F-Measure are commonly used in the information retrieval as performance measure [49]. Receiver Operating Characteristic (ROC) curve serves as graphical representation of the trade-off between the false negative and false positive rates for every possible cut off.

### 4.1 K-Fold Cross-Validation

To train and evaluate model statistical performance on the same data yields an overoptimistic result. Cross-validation was used to fix such a problem, starting from the remark that testing the output of the algorithm on new data would yield a good estimate of its performance accuracy [50]. Cross-validation is a statistical method that evaluates and compares machine learning schemes by dividing data into train and test set. The train set is used to learn or train a model and test set is used to validate the model. In practice, the training and validation sets must cross-over successively such that each data point has a chance of being validated against [51]. $k-$ Fold cross-validation is often used to minimize the bias associated with the random sampling of the training and hold-out data samples in comparing the baseline performance accuracy of two or more methods or classifiers [12].

In $k-$ fold cross-validation, the data is first partitioned into $k$ equally (or near equally) sized folds. The $k$ iterations are subsequently trained and validated such that within each iteration, a different fold of the data is held-out for validation while the remaining $k-1$ folds are used for learning. Prior to data splits into $k-$ folds, the data is stratified to rearrange the data so as to ensure each fold is a good representative of the entire datasets [51]. The folds are often stratified since cross-validation accurately depends on the random assignment of the individual cases into $k$ distinct folds. The folds are stratified in a manner that they contain approximately the same proportion of predictor labels as the original dataset [12] and [14]. In $k-$ fold cross-validation, the given dataset $S$ is randomly split into $k$ mutually exclusive subsets $(S_1, S_2, ..., S_k)$ of approximately equal size. The classifier is trained and tested $k$ times. Each time $t \in \{1, 2, ..., k\}$, it is trained on all but one fold $S_t$ and tested on the remaining single fold $S_t$. The cross-validation estimate of the overall accuracy is calculated simply as the average of the $k$ individual accuracy measures as shown in Eq. (5)

$$\text{CVA} = \sum_{i=1}^{k} A_i \qquad (5)$$

where CVA stands for cross-validation accuracy, k is the number of folds and A is the accuracy measure of each fold [14].

### 4.2. Accuracy, Sensitivity & Specificity

The efficiency of any machine learning model is determined using measures such as True Positive Rate, False Positive Rate, True Negative Rate and False Negative Rate [52]. The sensitivity and specificity measures ae commonly used to explain clinical diagnostic test, and to estimate how good and consistent was the diagnostic test [53]. The sensitivity metrics is the true positive rate or positive class accuracy, while specificity is referred to as true negative rate or negative class accuracy. However, there is often a trade-off between the four performance measure metrics in "real world" applications [54]

a) Accuracy – Eq. (6) compares how close a new test value is to a value predicted by if ... then rules [22].

$$Accuracy = \frac{TP+TN}{TP+TN+FP+FN} 100\% \quad (4)$$

b) Sensitivity – Eq. (7) measures the ability of a test to be positive when the condition is actually present. It is also known as false-negative rate, recall, Type II error, β error, error of omission, or alternative hypothesis [22].

$$Sensitivity = \frac{TP}{TP+FN} 100\% \quad (5)$$

c) Specificity – Eq. (8) measures the ability of a test to be negative when the condition is actually not present. It is also known as false-positive rate, precision, Type I error, α error, error of commission, or null hypothesis [22].

$$Specificity = \frac{TN}{TN+FP} 100\% \quad (6)$$

d) Predictive Accuracy – Eq. (9) gives an overall evaluation. It is also known as the percentage proportion of correctly classified cases to all cases in the set. The larger the predictive accuracy the better the situation.

$$PA = \frac{TP+TN}{TP+TN+FP+FN} 100\% \quad (7)$$

4.3. Recall, Precision and F-measure

Eqs. (10-12) are additional parameters that could help physician determine exactly whether a patient is ill or not. Recall is the same in application as sensitivity. F-measure is the harmonic mean of both recall and precision, while specificity is the reverse of sensitivity.

$$Precision = \frac{TP}{TP+FP} \quad (8)$$

$$Recall = \frac{TP}{TP+FN} \quad (9)$$

$$F\text{-measure} = \frac{2*precision*recall}{precision+recall} \quad (10)$$

4.4. Error Costs and Estimation

Some of the widely used errors variants incorporated into most machine learning tool(s) include [23].

a) Mean absolute error (MAE): Eq. (13) is the average of individual errors while neglecting the signs to diminish the negative effects of outliers [23].

$$\frac{|p_1-a_1|+...+|p_n-a_n|}{n} \quad (11)$$

b) Root mean square error

$$\sqrt{\frac{(p_1-a_1)^2+...+(p_n-a_n)^2}{n}} \quad (12)$$

c) Relative absolute error

$$\frac{|p_1-a_1|+...+|p_n-a_n|}{|a_1-\bar{a}|+...+|a_n-\bar{a}|} \quad (13)$$

where $\bar{a} = \frac{1}{n}\sum_i a_i$ denotes a total absolute error normalized by the error of a predictor which uses an average of the actual values from a dataset [23].

d) Root relative squared error:

$$\sqrt{\frac{(p_1-\bar{a})^2+...+(p_n-a_n)^2}{|a_1-\bar{a}|^2+...+|a_n-\bar{a}|^2}} \quad (14)$$

where $p$ is predicted target values $p_1, p_2,..., p_n$ while $a$ represents actual value: $a_1, a_2,..., a_n$

4.4 Receiver Operating Characteristics (ROC)

ROC summarizes classifier performance over a range of trade-offs between true positive $TP$ and false positive $FP$ error rates. The Areas under the Curve (AUC) is accepted performance metric for the Receiver Operating Characteristic (ROC) curve. On the ROC curve, the ROC plots the curve with $X-$ axis to represents percentage (%) of false positive $FP$ ; $\%(FP) = \frac{FP}{(TN+FP)}$ and plot the curve with $Y-$ axis to represents percentage (%) of true positive $TP$ ; $\%(TP) = \frac{TP}{(TP+FN)}$, and the ideal point on the ROC Curve would be an interval between (0,100) [27]. ROC is useful for exploring the trade-offs among different classifiers over a range of scenarios, which is not ideal for situations with known error costs. The area under the curve (AUC) is most preferred because the larger the area the better the model. The AUC also has a nice interpretation as the probability that the classifier ranks a randomly chosen positive instance above a randomly chosen negative one [55]. Area under the ROC Curve (AUC) is a useful metric for classifier performance as it is independent of the decision criterion selected and prior probabilities. AUC can establish a dominance relationship between classifiers. If the ROC curves are intersecting, the total AUC is an average comparison between models [27].

## 5. Empirical Results and Analysis

### 5.1 Thoracic Surgery Dataset

The thoracic surgery database was obtained from the University of California Irvine (UCI) machine learning repository database [56]. The thoracic datasets is dedicated to classification problem related to the post-operative life expectancy in the lung cancer patients: class 1 - death within one year after surgery, class 2 - survival; recoded as Risk1Y: 1 year survival period - (T)rue value if died (T, F), where the class value (Risk1Y) is binary valued. The thoracic surgery datasets are not approximately equally distributed. The data was collected retrospectively at Wroclaw Thoracic Surgery Centre for patients who underwent major lung resections for primary lung cancer for period of 4 years [57]. The dataset contains 470 instances, and 17 attributes, 14 of which are nominal and 3 numeric – Age, PRE4 and PRE5 as shown in Table 1.

Although random over-sampling can increase the likelihood of over-fitting occurring, it may introduce an additional computational task if the dataset is already fairly large but imbalanced. Synthetic Minority Over-sampling Technique (SMOTE) in WEKA was used to generate synthetic minority examples to over-sample the minority class. The aim is to form new minority class examples by interpolating between several minority classes examples that lie together, using the concept of k-nearest neighbour. In this way, the over fitting problem was avoided causing the decision boundaries for the minority class to spread further into the majority class space [58]. The predicted class Risk1Y: 1 year survival period - (T)rue value if died (T, F), where the class value (Risk1Y) is binary valued, initially had a sample distribution of T (70) and F (400). After repeated application of Synthetic Minority Over-sampling Technique (SMOTE) randomize in WEKA, the same sample distribution was now T (560) and F (400).

Table 1 Thoracic Surgery database

| Thoracic Dataset Recode | Attributes | |
|---|---|---|
| | Description | Values (Nominal/Numeric) |
| DGN | Diagnosis-specific combination of ICD-10 codes for primary and secondary as well multiple tumors if any | DGN3,DGN2,DGN4,DGN6,DGN5, DGN8,DGN1 |
| PRE4 | Forced vital capacity - FVC | Numeric |
| PRE5 | Volume that has been exhaled at the end of the first second of forced expiration - FEV1 | Numeric |
| PRE6 | Performance status - Zubrod scale | PRZ2,PRZ1,PRZ0 |
| PRE7 | Pain before surgery | T, F |
| PRE8 | Haemoptysis before surgery | T, F |
| PRE9 | Dyspnoea before surgery | T, F |
| PRE10 | Cough before surgery | T, F |
| PRE11 | Weakness before surgery | T, F |
| PRE14 | T in clinical TNM - size of the original tumor, from smallest to largest | OC11,OC14,OC12, OC13 |
| PRE17 | Type 2 DM - diabetes mellitus | T, F |
| PRE19 | MI up to 6 months | T, F |
| PRE25 | PAD - peripheral arterial diseases | T, F |
| PRE30 | Smoking | T, F |
| PRE32 | Asthma | T, F |
| AGE | Age at surgery | Numeric |
| Risk1Yr | 1 year survival period - (T)rue value if died (T,F) | T, F |

## 6. Results and Discussion

Waikato Environment for Knowledge Analysis (WEKA) was used to simulate the baseline performance accuracy of the classifiers in a more convenient manner to determine which scheme is statistically better than the other [59]. Stratified 10-fold cross-validation was used to rearrange the data to ensure that each fold is a good representation of the whole datasets. The stratified 10-fold cross-validation (k = 10) is the most common [60], and universal [50] evaluation models, with lower sample distribution variance compared to the hold-out cross-validation. Final analysis show that Multilayer Perceptron classifier performed best with classification accuracy of 82.3%, with True Positive rate of 82.4% and a ROC Area (AUC) of 84.7% with a minimum error rate of 21.6% (mean absolute error) and maximum error rate of 78.4% (root relative squared error). J48 classifier came out to be second best with a classification accuracy of 81.8%, with True Positive rate of 81.9% and a ROC Area (AUC) of 82.2% with a minimum error rate of 22.7% (mean absolute error) and maximum error rate of 80.6% (root relative squared error). The Naive Bayes classifier came out to be worst of the three algorithms with a classification accuracy of 74.4%, with True Positive rate of 74.5% and a ROC Area (AUC) of 79.2% with a minimum error rate of 29.0% (mean absolute error) and maximum error rate of 91.2% (root relative squared error). Table 2 and Fig. 3 present the detail results of the comparative analysis for better visualization and analysis.

Table 2 10-fold cross-validation performance evaluation comparison

| Performance Metrics | MLP | J48 | Naïve Bayes |
|---|---|---|---|
| Correctly Classified Instances | 82.3 | 81.8 | 74.4 |
| Mean absolute error | 21.6 | 22.7 | 29.0 |
| Root mean squared error | 38.6 | 39.7 | 44.9 |
| Relative absolute error | 44.5 | 46.7 | 59.7 |
| Root relative squared error | 78.4 | 80.6 | 91.2 |
| True Positive (TP) Rate | 82.4 | 81.9 | 74.5 |
| False Positive (FP) Rate | 20.7 | 20.1 | 33.2 |
| Precision | 82.5 | 81.8 | 76.8 |
| Recall | 82.4 | 81.9 | 74.5 |
| F-Measure | 82.1 | 81.8 | 72.7 |
| ROC Area (AUC) | 84.7 | 82.2 | 79.2 |

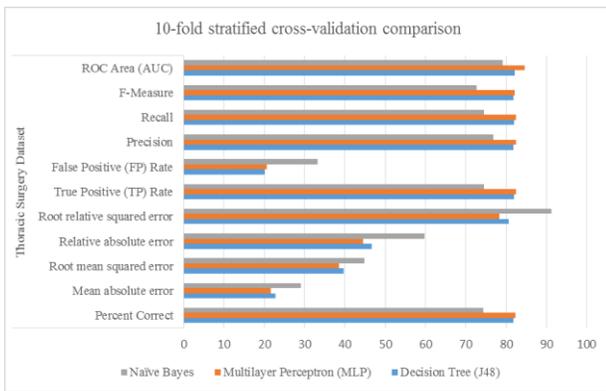

Fig.3 10-fold Stratified cross-validation performance comparison

## 7. Conclusions

This work was to identify and evaluate the performance of machine learning classification schemes applied in the prediction of post-operative life expectancy in Lung Cancer patients. Thoracic surgery dataset used for this study was obtained from the University of California Irvine (UCI) machine learning repository database. Data was collected retrospectively at Wroclaw Thoracic Surgery Centre for patients who underwent major lung resections for primary lung cancer for period of 4 years [57]. The dataset contained 470 instances, and 17 attributes, 14 of which are nominal and 3 numeric – Age, PRE4 and PRE5. The thoracic surgery datasets are not equally distributed. Synthetic Minority Over-sampling Technique (SMOTE) was used to generate synthetic minority examples to over-sample the minority class and smoothen the sample distribution from T (70) and F (400) to T (560) and F (400) after repeated application of Synthetic Minority Over-sampling Technique (SMOTE) randomize in WEKA.

SMOTE operates by interpolating between several minority classes examples that lie together, using the concept of k-nearest neighbor to avoid over fitting causing the decision boundaries for the minority class to spread further into the majority class space [58].

In this study, the thoracic datasets is dedicated to classification problem related to the post-operative life expectancy in the lung cancer patients: class 1 - death within one year after surgery, class 2 - survival; recoded as Risk1Y: 1 year survival period - (T)rue value if died (T, F), where the class value (Risk1Y) is binary valued. Multilayer Perceptron, J48 and Naive Bayes machine learning algorithms were calibrated to optimize the baseline performance accuracy of each classifier. Also stratified 10-fold cross-validation was used to measure the unbiased predictive accuracy of each classifier compared. Based on the stratified 10-fold cross-validation comparative analysis, Multilayer Perceptron achieved classification accuracy of 82.3%, true positive (TP) rate of 82.4%, and a ROC Area (AUC) of 84.7%. The J48 classifier achieved a classification accuracy of 81.8%, true positive (TP) rate of 81.9%, and a ROC Area (AUC) of 82.2%. And Naive Bayes classifier achieved a classification accuracy of 74.4%, true positive (TP) rate of 74.5%, and a ROC Area (AUC) of 79.2%.

Constraints involved in mining privacy-sensitive, heterogeneous and voluminous data must be considered in clinical mining. It is the responsibility of the researcher to anonymize data to ensure privacy-preserved clinical mining [21]. The clinical data contains either unique binary-valued attributes or multi-valued attributes from positive defined interval such as blood pressure or body temperature; and negative values resulting from screening diagnosis such as ECG, and RTG [23]. The uniqueness of clinical data requires researchers to take into consideration ethical, security and legal aspects of clinical mining, such as data ownership, fear of lawsuits, expected benefits, and special administrative issues [22]. However, the quality of machine learning algorithms applicable in clinical prognosis is dependent on the ability of the researcher to carefully choose, tune and apply machine learning classification to clinical prognosis.

### References

[1] M. J. Glenn, Making Sense of Data - A Practical Guide to Exploratory Data Analysis and Data Mining, Hoboken, New Jersey: John Wiley & Sons, Inc., 2007.
[2] U. M. Fayyad, G. Piatetsky-Shapiro, P. Smyth and R. Uthurusamy, Advances in Knowledge Discovery and Data Mining, Menlo Park, California: AAAI Press, 1996. A. A. Name, and B. B. Name, Book Title, Place: Press, Year.
[3] E. Kolce and N. Frasheri, "A Literature Review of Data Mining Techniques Used in Healthcare Databases," in ICT Innovations 2012, Web Proceedings, 2012.

[4] V. Patil and V. B. Nikam, "Study of Data Mining algorithm in cloud computing using MapReduce Framework," Journal of Engineering, Computers & Applied Sciences (JEC&AS), vol. 2, no. 7, pp. 65-70, July 2013.

[5] E. Osmanbegović and M. Suljić, "Data Mining Approach For Predicting Student Performance," Economic Review – Journal of Economics and Business, vol. X, no. 1 , pp. 3-12, May 2012.

[6] X. Geng and Z. Yang, "Data Mining in Cloud Computing," International Conference on Information Science and Computer Applications (ISCA 2013）, pp. 1-7, 2013.

[7] H. Chen, R. H. L. Chiang and V. C. Storey, "Business intelligence and analytics: From big data to big impact," MIS Quarterly, vol. 36, no. 4, pp. 1165-1188, December 2012.

[8] A. Guazzelli, "Predictive analytics in healthcare: The importance of open standards," IBM Corporation, 2011.

[9] R. Bellazzi and B. Zupan, "Predictive data mining in clinical medicine: Current issues and guidelines," International Journal of Medical Informatics, vol. 77, no. 2, pp. 81-97, February 2008.

[10] R. Bellazzi, F. Ferrazzi and L. Sacchi, "Predictive data mining in clinical medicine: a focus on selected methods and applications," WIREs Data Mining Knowledge and Discovery, vol. 1, no. 5, pp. 416-430, 11 February 2011.

[11] J. K. Pal, "Usefulness and applications of data mining in extracting information from different perspectives," Annals of Library and Information Studies, vol. 58, pp. 7-16, 2011.

[12] D. Delen, G. Walker and A. Kadam, "Predicting breast cancer survivability: a comparison of three data mining methods," Artificial Intelligence in Medicine, vol. 34, no. 2, pp. 113-127, 2005.

[13] K. Srinivas, B. K. Rani and A. Govrdhan, "Applications of Data Mining Techniques in Healthcare and Prediction of Heart Attacks," International Journal on Computer Science and Engineering (IJCSE), vol. 2, no. 2, pp. 250-255, 2010.

[14] D. Dursun, "Analysis of cancer data: a data mining approach," Expert Systems: The Journal of Knowledge Engineering, vol. 26, no. 1, pp. 100-112, February 2009.

[15] S. Floyd, "Data Mining Techniques for Prognosis in Pancreatic Cancer," 2007.

[16] T. Patrick, S. Laduzenski, D. Edwards, N. Ellis, A. P. Boezaart and H. Aygtug, "Use of machine learning theory to predict the need for femoral nerve block following ACL repair," Pain Medicine, vol. 12, no. 10, pp. 1566-1575, 2011.

[17] M.-L. Antonie, O. R. Zaı̈ane and A. Coman, "Application of Data Mining Techniques for Medical Image Classification," in Proceedings of the Second International Workshop on Multimedia Data Mining (MDM/KDD'2001), San Francisco, USA, 2001.

[18] J. P. Choi, T. H. Han and R. W. P. , "A Hybrid Bayesian Network Model for Predicting Breast Cancer Prognosis," Journal of Korean Society of Medical Informatics, vol. 15, no. 1, pp. 49-57, 2009.

[19] B. Abdelghani and E. Guven, "Predicting breast cancer survivability using data mining techniques," Ninth Workshop on Mining Scientific and Engineering Datasets in conjunction with the Sixth SIAM International, vol. 58, no. 13, pp. 10-110., 2006.

[20] M. U. Khan, J. P. Choi, H. Shin and M. Kim., "Predicting breast cancer survivability using fuzzy decision trees for personalized healthcare," in In Engineering in Medicine and Biology Society (EMBS) 2008. 30th Annual International IEEE EMBS Conference, Vancouver, British Columbia, Canada, 2008.

[21] J. J. Berman, "Confidentiality issues for medical data miners," Artificial Intelligence in Medicine, vol. 26, no. 1, p. 25–36, March 2002.

[22] K. J. Ciosa and G. W. Moore, "Uniqueness of medical data mining," Artificial Intelligence in Medicine, vol. 26, no. 1, p. 1–24, 2002.

[23] K. Aftarczuk, "Evaluation of selected data mining algorithms implemented in Medical Decision Support Systems," Blekinge Institute of Technology School of Engineering, Blekinge, 2007.

[24] J. Iavindrasana, G. Cohen, A. Depeursinge, H. Müller, R. Meyer and A. Geissbuhler, "Clinical Data Mining: a Review," Switzerland, 2009.

[25] J. Kaiser, "Dealing with Missing Values in Data," Journal of Systems Integration, vol. 5, no. 1, pp. 42-51, 2014.

[26] A. S. G. M. J. d. J. Fernández and F. Herrera, "A study of the behaviour of linguistic fuzzy rule based classification systems in the framework of imbalanced data-sets," Fuzzy Sets and Systems, vol. 159, no. 18, pp. 2378-2398, 2008.

[27] N. V. Chawla, K. W. Bowyer, L. O. Hall and W. P. Kegelmeyer, "SMOTE: Synthetic Minority Over-sampling Technique," Journal of Artificial Intelligence Research, p. 321–357, 2002.

[28] A. K. Menon, H. Narasimhan, S. Agarwal and S. Chawla, "On the Statistical Consistency of Algorithms for Binary Classification under Class Imbalance," In Proceedings of the 30th International Conference on Machine Learning, Atlanta, Georgia USA, 2013.

[29] E. Stamatatos, "Stamatatos, Efstathios."Author identification: Using text sampling to handle the class imbalance problem." Information Processing & Management," Information Processing & Management, vol. 44, no. 2, pp. 790-799, 2008.

[30] G. Cohen, M. Hilario, H. Sax, S. Hugonnet and A. Geissbuhler, "Learning from imbalanced data in surveillance of nosocomial infection," Artificial Intelligence in Medicine, vol. 37, no. 1, pp. 7-18, 2006.

[31] Y. Liu, X. Yu, J. X. Huang and A. Aijun, "Combining integrated sampling with SVM ensembles for learning from imbalanced datasets," Information Processing & Management, vol. 47, no. 4, pp. 617-631, 2011.

[32] M. E. Anbarasi, Anupriya and N. C. S. N. Iyengar, "Enhanced prediction of heart disease with feature subset selection using genetic algorithm," International Journal of Engineering Science and Technology, vol. 2, no. 10, pp. 5370-5376., 2010.

[33] J. Soni, U. Ansari, D. Sharma and S. Soni, "Predictive data mining for medical diagnosis: An overview of heart disease prediction," International Journal of Computer Applications, vol. 17, no. 8, pp. 43-48, 2011.

[34] M. Ture, F. Tokatli and I. Kurt, "Using Kaplan–Meier analysis together with decision tree methods (C&RT, CHAID, QUEST, C4. 5 and ID3) in determining recurrence-free survival of breast cancer patients." Expert Systems with Applications, vol. 36, no. 2, pp. 2017-2026, 2009.

[35] B.Venkatalakshmi and M. Shivsankar, "Heart Disease Diagnosis Using Predictive Data mining," in 2014 IEEE

International Conference on Innovations in Engineering and Technology (ICIET'14), Tamil Nadu, India, 2014.

[36] D. Van Den Poel and C. Kristof, "Churn prediction in subscription services: An application of support vector machines while comparing two parameter-selection techniques," Expert Systems with Applications, vol. 34, no. 1, p. 313–327, 2008.

[37] B. Milovic and M. Milovic, "Prediction and Decision Making in Health Care using Data Mining," International Journal of Public Health Science (IJPHS), vol. 1, no. 2, pp. 69-78, December 2012.

[38] M. Hemalatha and S. Megala, "Mining Techniques in Heath Care: A Survey of Immunization," Journal of Theoretical and Applied Information Technology, vol. 25, no. 2, pp. 65-70, 2011.

[39] B. K. Bhardwaj and S. Pal, "Data Mining: A prediction for performance improvement using classification," International Journal of Computer Science and Information Security (IJCSIS), vol. 9, no. 4, pp. 1-5, 2011.

[40] N. Chaudhary, Y. Aggarwal and R. K. Sinha, "Artificial Neural Network based Classification of Neurodegenerative Diseases," Advances in Biomedical Engineering Research (ABER), vol. 1, no. 1, pp. 1-8, 2013.

[41] A. S. Fathima, D. Manimegalai and N. Hundewale, "A Review of Data Mining Classification Techniques Applied for Diagnosis and Prognosis of the Arbovirus-Dengue," International Journal of Computer Science Issues (IJCSI), vol. 8, no. 6, pp. 322-328, November 2011.

[42] Q. K. Al-Shayea, "Artificial Neural Networks in Medical Diagnosis," International Journal of Computer Science Issues (IJCSI), vol. 8, no. 2, pp. 150-154, 2011.

[43] I. S. H. Bahia and Q. K. Al-Shayea, "Urinary System Diseases Diagnosis Using Artificial Neural Networks," IJCSNS International Journal of Computer Science and Network Security, vol. 10, no. 7, pp. 118-122, 2010.

[44] F. Amato, A. López, E. M. Peña-Méndez, P. Vaňhara, A. Hampl and J. Havel, "Artificial neural networks in medical diagnosis," Journal of Applied Biomedicine, vol. 11, no. 2, p. 47–58, 2013.

[45] S. Gupta, D. Kumar and A. Sharma, "Data mining classification techniques applied for breast cancer diagnosis and prognosis," Indian Journal of Computer Science and Engineering, vol. 2, no. 2, pp. 188-193, 2011.

[46] M. Kumari and S. Godara, "Comparative Study of Data Mining Classification Methods in Cardiovascular Disease Prediction," International Journal of Computer Science and Technology (IJCST), vol. 2, no. 2, pp. 304-308, June 2011.

[47] I. Y. Khan, P. Zope and S. Suralkar, "Importance of Artificial Neural Network in Medical Diagnosis disease like acute nephritis disease and heart disease," International Journal of Engineering Science and Innovative Technology (IJESIT), vol. 2, no. 2, pp. 210-217, 2013.

[48] X. Li, G. C. Nsofor and L. Song, "A comparative analysis of predictive data mining techniques," International Journal of Rapid Manufacturing, vol. 1, no. 2, pp. 50-172, 2009.

[49] C. Chao, A. Liaw and L. Breiman, "Using random forests to learn imbalanced data," University of California, Berkeley, 2004.

[50] S. Arlot and A. Celisse, "A survey of cross-validation procedures for model selection," Statistics Surveys, vol. 4, pp. 40-79, 2010.

[51] P. Refaeilzadeh, L. Tang and H. Liu, "Cross-validation," in In Encyclopedia of database systems, 2009.

[52] S. Deepa and V. Bharathi, "Textural Feature Extraction and Classification of Mammogram Images using CCCM and PNN," IOSR Journal of Computer Engineering (IOSR-JCE), vol. 10, no. 6, pp. 07-13, June 2013.

[53] R. Chitra and V. Seenivasagam, "Heart Disease Prediction System Using Supervised Learning Classifier," Bonfring International Journal of Software Engineering and Soft Computing, vol. 3, no. 1, pp. 1-7, March 2013.

[54] M. Agrawal, G. Singh and R. K. Gupta, "Predictive Data Mining for Highly Imbalanced Classification," International Journal of Emerging Technology and advanced Engineering, vol. 2, no. 12, pp. 139-143, 2012.

[55] H. W. Ian, E. Frank and M. A. Hall, Data Mining-Practical Machine Learning Tools and Techniques, 3rd ed., Burlington, MA USA: Morgan Kaufmann - Elsevier, 2011.

[56] K. Bache and M. Lichman, "UCI Machine Learning Repository," 2013.

[57] M. Ziaba, J. M. Tomczak, M. Lubicz and J. Swiatek, "Boosted SVM for extracting rules from imbalanced data in application to prediction of the post-operative life expectancy in the lung cancer patients," Applied Soft Computing, vol. 14, pp. 99-108, 2014.

[58] S. Kotsiantis, D. Kanellopoulos and P. Pintelas, "Handling imbalanced datasets: A review," GESTS International Transactions on Computer Science and Engineering, vol. 30, no. 1, pp. 25-36, 2006.

[59] R. R. Bouckaert, E. Frank, M. Hall, R. Kirkby, P. Reutemann, A. Seewald and D. Scuse, "Waikato Environment for Knowledge Analysis (WEKA) Manual for Version 3-7-8," The University of Waikato, Hamilton, New Zealand, 2013.

[60] L. T. H. L. Payam Refaeilzadeh, "Cross-Validation," in In Encyclopedia of database systems, US, 2009.